# BTMuda: A Bi-level Multi-source unsupervised domain adaptation framework for breast cancer diagnosis

Yuxiang Yang, Xinyi Zeng, Pinxian Zeng, Binyu Yan, Xi Wu, Jiliu Zhou, Yan Wang, *Member, IEEE*

*Abstract*—Deep learning has revolutionized the early detection of breast cancer, resulting in a significant decrease in mortality rates. However, difficulties in obtaining annotations and huge variations in distribution between training sets and real scenes have limited their clinical applications. To address these limitations, unsupervised domain adaptation (UDA) methods have been used to transfer knowledge from one labeled source domain to the unlabeled target domain, yet these approaches suffer from severe domain shift issues and often ignore the potential benefits of leveraging multiple relevant sources in practical applications. To address these limitations, in this work, we construct a Three-Branch Mixed extractor and propose a Bi-level Multi-source unsupervised domain adaptation method called BTMuda for breast cancer diagnosis. Our method addresses the problems of domain shift by dividing domain shift issues into two levels: intra-domain and inter-domain. To reduce the intra-domain shift, we jointly train a CNN and a Transformer as two paths of a domain mixed feature extractor to obtain robust representations rich in both low-level local and high-level global information. As for the inter-domain shift, we redesign the Transformer delicately to a three-branch architecture with cross-attention and distillation, which learns domain-invariant representations from multiple domains. Besides, we introduce two alignment modules - one for feature alignment and one for classifier alignment - to improve the alignment process. Extensive experiments conducted on three public mammographic datasets demonstrate that our BTMuda outperforms state-of-the-art methods.

*Index Terms*—Multi-source unsupervised domain adaptation, Three-Branch Transformer, Breast cancer screening, Intra-domain and Inter-domain Shift.

## I. INTRODUCTION

Breast cancer is the leading cause of cancer death in women worldwide, making early diagnosis crucial for decreasing mortality rates and saving lives [1]. In the clinic, it can be challenging for physicians to accurately diagnose whether a patient is normal or has a benign or malignant breast tumor based on major abnormal signs (e.g., masses, calcifications, architectural distortion, asymmetric dense shadowing) visible on X-ray scans [2]. With the development of deep learning (DL), great progress has been made in the automatic diagnosis of medical diseases [3-8]. However, achieving such success often requires the availability of large-scale and high-quality labeled data for training, which can be extremely time-consuming and labor-intensive in practice. To address this, some approaches [9, 10] have considered transfer learning, where they pre-trained the model using existing public datasets like ImageNet and then fine-tuned the model using a few labeled target data. While this strategy improves the performance of DL-based methods compared to training from scratch, the performance gain is often limited by the significant difference in data distribution between natural and medical image domains, known as domain shift. Besides, the potential of the existing large number of unlabeled data has not been fully exploited.

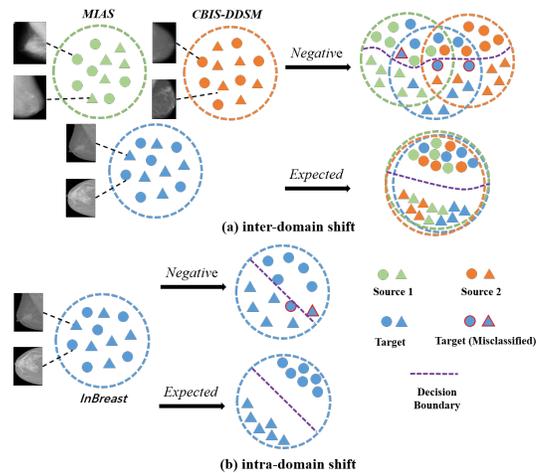

**Fig. 1.** Illustration of the Bi-level domain shifts. (a) inter-domain shift. (b) intra-domain shift.

Recently, unsupervised domain adaptation (UDA), a challenging branch of transfer learning, has gained wide research attention due to its ability to reduce the expensive cost of labeling data. UDA enables the knowledge to be transferred from a labeled source domain to a relevant unlabeled target domain by alleviating the domain shift

This work is supported by National Natural Science Foundation of China (NSFC 62371325, 62071314), Sichuan Science and Technology Program 2023YFG0025, 2023YFG0101, and 2023 Science and Technology Project of Sichuan Health Commission 23LCYJ002.

This work did not involve human subjects or animals in its research.

Yuxiang Yang, Xinyi Zeng, Pinxian Zeng, Binyu Yan, Jiliu Zhou and Yan Wang are with the School of Computer Science, Sichuan University, Chengdu 610065, China. (e-mail: yangyuxiang3@stu.scu.edu.cn; perperstudy@gmail.com; 651215874@qq.com; yanbinyu2023@163.com; zhoujiliu@cuit.edu.cn; wangyanscu@hotmail.com).

Xi Wu is with School of Computer Science, Chengdu University of Information Technology, Chengdu 610103, China (e-mail: wuxi@cuit.edu.cn).

between them. For instance, Wang *et al*. [11] proposed an adversarial learning method with a domain adaptation stage and a case-level learning stage to improve the performance of breast cancer diagnosis using UDA. However, most existing works [11-14] in UDA focus on using only a single source domain (Single-source Unsupervised Domain Adaptation, SUDA). In practice, labeled source data can be collected from multiple source domains (e.g., multiple clinical centers), which may contain more comprehensive and discriminative knowledge that could be beneficial for the model in the target domain. In light of this, Multi-source Unsupervised Domain Adaptation (MUDA), a powerful extension of UDA that leverages multiple source domains, has attracted increasing attention from researchers [15-20]. Nevertheless, few works have considered the MUDA setting in the context of breast cancer diagnosis or even across the entire field of medical image analysis. We argue that the main reason can be attributed to the following two challenges. First, medical datasets collected by different clinical centers or institutions tend to have diverse characteristics due to the differences in data collection procedure, equipment, and experimental setting. This heterogeneity can lead to greater inter-domain shifts between any two domains in the MUDA setting [21], making it more difficult to align multiple domains in the common feature space as expected. As shown in Fig. 1 (a), the existing inter-domain shift can negatively impact the model's ability to learn domain-invariant representations and cluster data with the same class but in different domains. Therefore, the model may learn a misaligned feature space with a distorted decision boundary that does not generalize well to the target domain, leading to misclassified predictions. Second, data collected from the same clinical center or even from the same equipment could also have diverse distributions caused by various factors, such as shooting angle, individual differences, etc. This distribution difference within the dataset, known as intra-domain shift, still hinders the performance in the target domain. As shown in Fig. 1 (b), the intra-domain shift can make it difficult for the model to learn a comprehensive and robust feature space and distinguish between different samples within the same domain, potentially resulting in the unlabeled target samples being distributed one-sidedly near the decision boundary and leading to incorrect predictions. These two challenges hinder in-depth analysis in existing studies and there is still great potential for studying MUDA in the field of medical image analysis.

To address the inter-domain shift issue, it is essential to mine sufficient inter-domain invariant representations (i.e., common features shared across different domains) during the domain alignment process. For instance, in a MUDA task with real human images and synthetic human images as the source domain, the model needs should extract the features that are more associated with shared category information, such as the overall human silhouette and human joints. In light of this, some works have attempted to align multiple sources and the target domain into a common feature space to learn inter-domain invariant representations by optimizing the consistency loss or adversarial loss between the predictions of source-target pairs [18, 22]. However, solely relying on output-level constraints, is not sufficient to address the inter-domain shift problem. Inspired by multi-modal networks [23, 24] that employ cross-attention for feature fusion, we propose to use the cross-attention strategy to build a fine-grained feature alignment process to generate a proper intermediate feature space among the source and target domains, thus relieving the domain shift from the feature-level perspective. Particularly, cross-attention allows the model to weigh the importance of tokens in one input (i.e., source domain) concerning tokens in another input (i.e., target domain), thus dynamically deriving an intermediate feature that is properly distributed between the target and source domains rather than simply performing a weighted sum. Meanwhile, by iteratively conducting cross-attention between features from each source and target domain, the model can continuously mine more inter-domain invariant representations from each source domain to facilitate knowledge transfer.

Compared to inter-domain shifts, the issue of intra-domain shifts in MUDA tasks has received less consideration. While there are a few methods that address this issue through specific intra-domain alignment procedures [22, 25], they usually depend solely on CNNs for feature extraction, which excels at extracting rich local information but struggles to capture long-range dependencies. Such a biased feature extraction process can result in a weakened feature space and one-sided distributions of target data, making it challenging to handle the intra-domain shift. Recently, Vision Transformer (ViT) [26] and its variants [27, 28] have made significant progress in various vision tasks due to their ability to capture long-range relationships and compute attention weights dynamically using the global self-attention mechanism. In light of this, instead of building a specialized module for intra-domain alignment, we propose a more fine-grained feature extraction process by simply combining Transformer and CNN to comprehensively extract representations with both global and local senses. This allows for the learning of rich discriminative category information and the mitigation of the intra-domain shift issue. Additionally, the aforementioned cross-attention, which addresses the inter-domain shift, can be conveniently integrated with self-attention as separate calculation branches in the Transformer path, enabling an effective derivation of both intermediate (calculated by cross-attention) and target/source features (calculated by self-attention) in a one-time forward process.

To sum up, in this paper, we construct a **T**hree-Branch Mixed extractor and propose a novel **B**i-level **M**ulti-source unsupervised domain adaptation method called BTMuda to simultaneously address both inter- and intra-domain shifts for breast cancer diagnosis (classification tasks) using multiple 2D mammography sources. Our main contributions are as follows:

1) We propose a novel solution for addressing both insufficient data and bi-level domain shifts in the multi-source domain adaptation task for breast cancer diagnosis.

2) We create a domain mixed feature extractor by co-training a CNN and a Transformer jointly to extract both the

local features as well as the global structural relations. This allows the model to have a more nuanced and comprehensive understanding of each domain, thus mitigating the intra-domain shift issue.

3) By introducing the cross-attention, we delicately redesign the Transformer into a three-branch architecture, which fuses inter-domain invariant representations from the source domains and the target domain to reduce the inter-domain shift problem. In addition, we also introduce two types of alignment modules for features and classifiers to further fine-tune the inter-domain alignment process.

4) Substantial experiments in three public mammographic datasets demonstrate that our method achieves state-of-the-art performance compared with the current MUDA methods.

## II. RELATED WORKS

### A. Single-Source Unsupervised Domain Adaptation

Single-source unsupervised domain adaptation (SUDA) aims to transfer knowledge from a labeled source domain to an unlabeled target domain. Current deep SUDA methods are mainly divided into two categories: instance-based and feature-based. The instance-based methods [29, 30] aim to align distributions of the source and target domains at the image level using generative adversarial networks (GANs) [31], while feature-based methods [12, 13] aim to minimize the discrepancy between the two domains at the feature level. In the field of medical image analysis, SUDA has attracted increasing attention due to its advantage of not requiring labeled target data, and several attempts have been made [11, 14]. In [11], an adversarial learning method was proposed to improve the performance of breast cancer screening using mammographic images. While these methods are effective for SUDA, they are not competitive for addressing more complex data distribution in MUDA.

### B. Multi-Source Unsupervised Domain Adaptation

Recently, MUDA has garnered extensive attention, as it deals with a more practical scenario in which labeled training samples are collected from multiple sources. There have been various approaches proposed for MUDA tasks on natural images. For example, Peng *et al.* [18] minimized the first-order moment-related distance between all source and target domains to learn domain invariant feature representations. Li *et al* [32] developed a feature filtration mechanism and designed a corresponding network to achieve a selective feature alignment based on the transferability of features.

While most MUDA research has focused on natural images, there have been a few studies exploring MUDA tasks in medical image analysis. Zhang *et al* [33] proposed a multi-source domain adaptation method for mitotic cell detection, aiming at extracting the candidate regions of mitosis and classifying them as mitosis or non-mitosis. Abbet *et al* [34] incorporated domain discrimination and image reconstruction into a meta-learning framework for prostate MRI segmentation. Li *et al* [15] integrated multiple biological information sources and constructs a dual-layer heterogeneous network for prognostic biomarkers of breast cancer. However, these previous works mainly focused on adapting models from source domains to the target domain to reduce the inter-domain shift, while the problem of intra-domain shift has received little attention.

### C. Convolutional Neural Networks and Vision Transformer

Convolutional neural networks (CNNs) have been widely used in the field of deep learning and have achieved successful results in various vision tasks [5, 7, 15]. Thanks to progressively expanded receptive fields, CNNs are particularly effective in extracting local information from hierarchies of structured image representations. On the other hand, transformers [35], which were originally developed for natural language processing (NLP), have also demonstrated record-breaking performance on various language tasks due to their ability to capture long-range dependencies through the attention mechanism. Inspired by the astounding performance, research has moved towards applying the same principles in computer vision (CV). For example, some studies attempted [36, 37] to integrate attention into CNNs to model heterogeneous interactions, and some works [38, 39] adopted a hybrid transformer and CNN architecture or directly used convolution-free architectures, such as the Vision Transformer (ViT) and its variants to improve the ability of the model.

## III. METHODOLOGY

In our MUDA setting, there are $M$ labeled source domains $S = \{S_1, ..., S_M\}$ and one unlabeled target domain $\mathcal{T}$. All the data of the source and target domains are collected from $M+1$ clinical centers and share the same category space. We expect to design a model that can adapt from multiple source datasets $\mathcal{D}^{S_M} = \{x_i^{S_M}, y_i^{S_M}\}_{i=1}^{N_{S_M}}$ in a source domains $S_M$ to the target dataset $\mathcal{D}^{\mathcal{T}} = \{x_i^{\mathcal{T}}\}_{i=1}^{N_{\mathcal{T}}}$ in the target domain $\mathcal{T}$. Here, $x$ is the input mammography image, $y = \{1, ..., C\}$ is the corresponding one-hot label with $C$ classes, $N_{S_M}$ and $N_{\mathcal{T}}$ are the numbers of samples in $S_M$ and $\mathcal{T}$. For breast cancer diagnosis, the diagnosis result of any sample could be positive (normal/benign) or negative (malignant), i.e., $C=2$. The goal of our work is to correctly classify target domain samples utilizing all the labeled source data and unlabeled target data.

The overall framework is illustrated in Fig. 2. The proposed BTMuda begins with a two-path domain mixed feature extractor, including a CNN-based feature extractor and a Transformer-based feature extractor running in parallel. These two feature extractors are co-trained with a consistency constraint to mine both local and global features, enabling the model to determine the data distribution of each category more comprehensively within each domain with a relieved intra-domain shift. Notably, the Transformer-based feature extractor is implemented with three branches and the cross-attention mechanism to alleviate the inter-domain shift between source and target domains. After the feature extractor along each path, we have two types of alignment modules that help to close the domain gap from different perspectives. The first module is the source-target feature alignment module consisting of $M$ alignment sub-modules, one for each source domain. These sub-modules can facilitate the domain mixed extractor to learn inter-domain invariant representations by aligning features from the target domain and each source domain separately. The other module is the decision boundary alignment module, which is made up of $M$ classifiers and uses a restriction loss to constrain all these classifiers to maintain consistent predictions for every sample in the target domain. To make full use of the heterogeneous features extracted from the CNN- and Transformer-based paths, a feature fusion module is added between two paths to fuse the CNN and Transformer features, encouraging the model to predict the diagnosis results more robustly. We will detailedly introduce these modules in the following sub-sections.

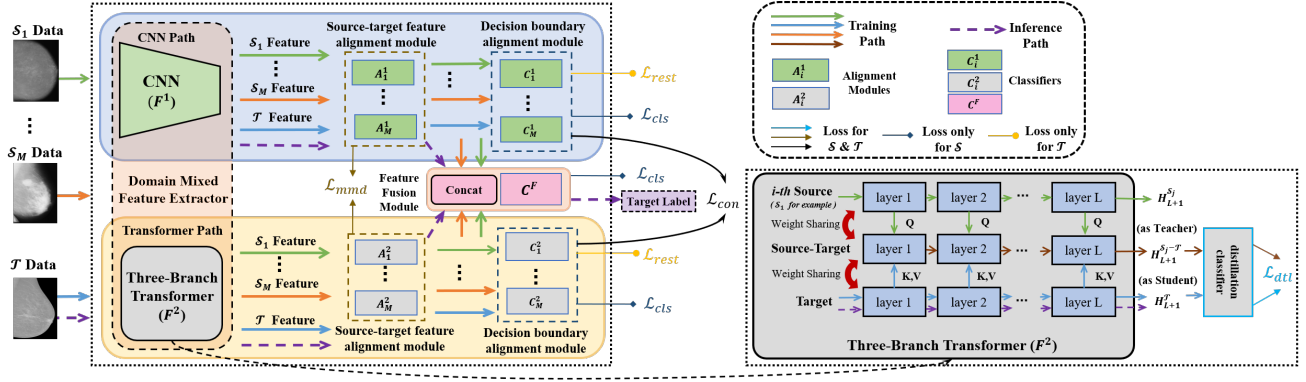

**Fig. 2.** Overview framework of our BTMuda. The modules marked in green belong to the CNN path, the modules marked in gray belong to the Transformer path, and the modules marked in pink belong to the feature fusion module; The purple dotted line ending with an arrow represents the data flow of the Target domain in the inference stage, while the other solid lines ending with an arrow represent the data flows of target and source domains in the training stage.

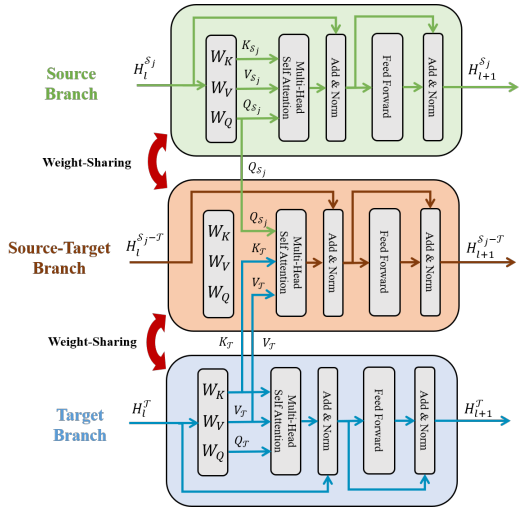

**Fig. 3.** Overview of the proposed three-branch Transformer (the $l$-th layer).

*A. Domain Mixed Feature Extractor*

As shown in Fig. 2, a CNN feature extractor $F^1$ is co-training with a three-branch Transformer $F^2$, serving as two extracting paths for a domain mixed feature extractor. Both $F^1$ and $F^2$ will receive paired mammography images $x_i^{S_j}$ and $x_i^T$ from the $j$-th source domain and the target domain to jointly obtain both local semantic information and global structural information. For the Transformer feature extractor $F^2$, it will be used to align each source-target pair by employing cross-attention in the three branches. Next, we will elaborate on the details of the Transformer feature extractor.

**Three-branch Cross-attention in the Transformer Path:** Our Transformer feature extractor $F^2$ consists of three branches, as shown in Fig. 3. In the training phase, the source and target domains use upper and lower branches, applying self-attention to learn discriminative representations. Concretely, a pair of images including a source image $x^{S_j}$ from the $j$-$th$ source domain and a target image $x^T$ is fed into the source and target branches, respectively, producing transformed patches $H_l^{S_j}$ and $H_l^T$ in the $l$-th layer (we omit the image index $i$ for simplicity). These patches are then projected into three vectors for the subsequent self-attention operation, namely query vectors $Q_{S_j}, Q_T \in R^{N*d_k}$, key vectors $K_{S_j}, K_T \in R^{N*d_k}$ and value vectors $V_{S_j}, V_T \in R^{N*d_v}$, where $d_k$ and $d_v$ are the dimension of the corresponding vectors and $N$ is the number of patches. The output of the self-attention operation is computed as a weighted sum of values, where the weight of each value is determined by the dot product of the corresponding query and key, followed by an activation function. This process can be written as:

$$Atten_{self}(x^{S_j}) = softmax\left(\frac{Q_{S_j}*K_{S_j}'}{\sqrt{d_k}}\right)*V_{S_j},$$
$$Atten_{self}(x^T) = softmax\left(\frac{Q_T*K_T'}{\sqrt{d_k}}\right)*V_T. \quad (1)$$

The self-attention mechanism in the source and target branches helps the network extract long-range, global representations, and the weight-sharing mapping, which generates the $Q$, $K$, and $V$ vectors for both branches, also provides the network with a certain level of generalization ability for the target data. However, since there is no interaction and alignment process between the features in the two branches, the network may still focus on learning error-prone domain-specific features and cannot effectively mine sufficient inter-domain invariant features to solve the severe inter-domain shift problem. To address this issue, we intuitively construct a source-target branch with a convenient cross-attention calculation. In this branch, the above $Q_{S_j}$ from the source branch as well as the $K_T$ and $V_T$ from the target branch will perform the following cross-attention calculation:

$$Atten_{cross}(x^{S_j}, x^T) = softmax\left(\frac{Q_{S_j} * K_T{'}}{\sqrt{d_k}}\right) * V_T. \quad (2)$$

In the calculation of cross-attention, the weight coefficient is determined by the similarity between the corresponding query vector and all the key vectors, with larger similarity resulting in a larger attention weight coefficient. Therefore, among all image patches in the target domain, patches that are more similar to the query vector in source domains are more likely to contain shared information and will be assigned higher weights. In this way, the model can mine more domain-invariant representations by measuring the similarity between patches in source/target domains. Notably, the output of the $(l-1)$-th layer in the source-target branch, denoted as $H_l^{S_j-T}$ and only exists when $l \geq 2$, will be then added to the output of the cross-attention module. After going through a feed-forward network, followed by an addition and normalization operation, the $l$-th layer of the three-branch Transformer will generate the inputs corresponding to each branch of the next layer, identified by symbols as $H_{l+1}^{S_j}$, $H_{l+1}^{S_j-T}$, and $H_{l+1}^{T}$, respectively.

**Two-Branch Distillation in the Transformer Path:** The above cross-attention mechanism in the source-target branch enables the model to dynamically mine inter-domain invariant representations from multiple source domains and the target domain during the intermediate forward process. In this subsection, we further exploit the source-target branch by utilizing its final fused features, which can also be seen as features from a potential intermediate domain. These fused features, which contain rich domain-invariant representations from both the source and target domains, are believed to have a relatively small inter-domain shift between the target domain compared with the original source domains. Therefore, we use these features to provide additional guidance for the learning of the target branch by treating the source-target domain branch as a "teacher" and the target branch as a "student". Specifically, after going through $L$ layers in the Transformer path, the outputs of the source-target branch $H_{l+1}^{S_j-T}$ and target branch $H_{l+1}^{T}$ will be fed into a unique distillation classifier $C^d$ to obtain the corresponding probabilities $P_{S_j-T}^d$, $P_T^d$, respectively. $P_{S_j-T}^d$ is then used as a soft label to supervise $P_T^d$ by a distillation loss [40], which is defined as:

$$\mathcal{L}_{dtl} = \frac{1}{M}\sum_{i=1}^{M} -softmax\left(P_{S_j-T}^d\right) * \log P_T^d. \quad (3)$$

**Co-training Constraints on two Paths:** To enhance the consistency between the predictions of the CNN and the Transformer feature extractors, we use a consistency loss to constrain the output predictions of the above CNN and Transformer paths. Herein, the predictions of the source data on each path are $P_{S_j}^k$ ($k = 1$ for CNN path and $k = 2$ for Transformer path), where we omit their subscripts as $P^k$ for simplicity since they undergo the same operation regardless of which domains they come from. The consistency loss can be formulated as follows:

$$\mathcal{L}_{con} = \frac{1}{M+1}\sum_{i=1}^{m+1}\frac{D_{KL}(\sigma(P^1)||\sigma(P^2)) + D_{KL}(\sigma(P^2)||\sigma(P^1))}{2}, \quad (4)$$

where $D_{KL}(\cdot)$ refers to the K-L distance, $P^1$ and $P^2$ represent final prediction probabilities for $x$ of the CNN branch and the Transformer branch, respectively. By enforcing the consensus between the two predictions, the domain mixed feature extractor can perceive both local and global information and a more nuanced and comprehensive understanding of each domain. Meanwhile, the final prediction can be verified from both local and global views, further alleviating the intra-domain shift and enhancing the robustness of the model.

*B. Alignment Modules*

**Source-target Pairs Alignment Module:** As the number of source domains increases, it is becoming more difficult to map all domains into the same feature space. Therefore, the source-target pairs alignment module is proposed as a fine-grained feature alignment process to learn domain-invariant representations and alleviate the inter-domain shift. In this module, the feature extracted by the previous module of the source domain and target domain is separately mapped to another corresponding feature space and then aligned by using the MMD loss [40], which is calculated as follows:

$$\mathcal{L}_{mmd} = \frac{1}{2M}\sum_{j=1}^{M}\sum_{k=1}^{2}\widehat{D}\left[A_j^k\left(F^k(x^{S_j})\right), A_j^k\left(F^k(x^T)\right)\right], \quad (5)$$

where $\widehat{D}$ represents the Reproduction Kernel Hilbert Space.

**Decision Boundary Alignment Module:** The source-specific classifiers in $\{C_j^k\}_{k=1,j=1}^{2*M}$ tend to generate different predictions for target samples that are adjacent to the decision boundary. This potentially leads the model to make unreliable predictions, as it may rely solely on knowledge learned from a closer source domain. To address this issue, we propose a decision boundary alignment module that minimizes differences among all classifiers in two paths $\{C_j^k\}_{k=1,j=1}^{2*M}$ and ensures that they provide the same predictions for every target domain sample. Specifically, the restriction loss will calculate the absolute value of the differences between the prediction probabilities of any classifier and other $(2M - 1)$ classifiers, thus resulting in a total of $C_{2M}^2 = M * (2M - 1)$ pairs. By minimizing this loss, the decision boundaries produced by each source domain classifier on the target domain are aligned, thus generating a more robust and comprehensive decision boundary for the prediction of the target data. The loss function for this operation is shown as follows:

$$P_{(k-1)*M+j} = C_j^k\left(A_j^k(F^k(x^T))\right), \quad (6)$$

$$\mathcal{L}_{rest} = \frac{1}{M*(2M-1)}\sum_{m=1}^{2M-1}\sum_{n=m+1}^{2M}|P_m - P_n|. \quad (7)$$

where we omit path index $k$ and source index $j$ in Eq. 7 by adopting the simplified expression in Eq. 6, which expresses the prediction probability of target domain data as $P_m$ and $P_n$ containing only subscripts.

*C. Feature Fusion Module and Classification Loss*

To make better use of features derived from the domain mixed CNN-Transformer extractor, we add a feature fusion module at the end. The feature fusion module concatenates the

**Algorithm 1:** Training procedure of our BTMuda.

1: **Input:** $M$ source datasets $\mathcal{D}^{S_M} = \{x_i^{S_M}, y_i^{S_M}\}_{i=1}^{N_{S_M}}$ with their corresponding labels and one single target dataset $\mathcal{D}^{\mathcal{T}} = \{x_i^{\mathcal{T}}\}_{i=1}^{N_{\mathcal{T}}}$ without labels.
2: **Initialize:** Initialize the network parameters: $\theta_{F^k}$ for $F^k$, $\theta_{A_i^k}$ for $A_i^k$, $\theta_{C_i^k}$ for $C_i^k$, $\theta_{C^d}$ for $C^d$, $\theta_{C^f}$ for $C^f$, $k \in \{1,2\}$, $i \in \{1, ..., M\}$.
3: **for** $e = 0$ to $iter\_total$ **do**
4:    Take out a batch of samples $B^{S_M}$ and $B^{\mathcal{T}}$ from $\mathcal{D}^{S_M}$ and $\mathcal{D}^{\mathcal{T}}$
5:    Compute alignment features $A_i^1(F^1(x_i^{S_M}))$ and $A_i^1(F^1(x_i^{\mathcal{T}}))$ from CNN path
6:    Compute common features $F^2(x_i^{S_M})$, $F^2(x_i^{\mathcal{T}})$ and $F^2(x_i^{S_M-\mathcal{T}})$ with Eq. (1) and Eq. (2) from Three-Branch Transformer
7:    Compute $C^d(F^2(x_i^{\mathcal{T}}))$ and $C^d(F^2(x_i^{S_M-\mathcal{T}}))$ to calculate the the distillation loss $\mathcal{L}_{dtl}$ with Eq. (3)
8:    Compute alignment features $A_i^2(F^2(x_i^{S_M}))$ and $A_i^2(F^2(x_i^{\mathcal{T}}))$ from Transformer path
9:    Calculate the alignment MMD loss $\mathcal{L}_{mmd}$ with Eq. (5)
10:   Compute source predictions $C_i^k(A_i^k(F^k(x_i^{S_M})))$ and target predictions $C_i^k(A_i^k(F^k(x_i^{\mathcal{T}})))$, $k \in \{1,2\}$
11:   Calculate the restriction loss $\mathcal{L}_{rest}$ with Eq. (7)
12:   Calculate the consistency loss $\mathcal{L}_{con}$ with Eq. (4)
13:   Compute the fusion source features and predictions with Eq. (8)
14:   Calculate the final source classification loss $\mathcal{L}_{cls}$ with Eq. (9)
15:   Update $\{\theta_{F^k}, \theta_{A_i^k}, \theta_{C_i^k}, \theta_{C^d}, \theta_{C^f}\}$ by minimizing Eq. (10)
16: **end for**

features obtained from the two paths of the domain mixed feature extractor into a single aggregated feature, which is then input to the embedded fusion classifier $C^f$ to produce a final prediction with more intra-domain robustness. The formulas for the prediction of the source domain data from the two paths as well as the feature fusion module are expressed as follows:

$$P_{S_j}^k = C_j^k\left(A_j^k\left(F^k(x^{S_j})\right)\right),$$
$$P_S^f = C^f\left(Cat_{j=1...M; k=1,2}\left(A_j^k\left(F^k(x^{S_j})\right)\right)\right), \quad (8)$$

where $P_{S_j}^k$ is the prediction probability generated by the classifier for the $j$-th source domain using the $k$-architecture in the two paths of the domain mixed feature extractor. As for $P_S^f$, no matter which source domain the data $x^{S_j}$ comes from, it will also pass through all alignment sub-modules, including those of other source domains to obtain all latent features. This enables the feature fusion module to learn knowledge from multiple source domains and generalize to unlabeled target data as much as possible to achieve a more robust inference process later. The final classification loss is defined as follows:

$$\mathcal{L}_{cls} = \frac{1}{2M}\sum_{j=1}^{M}\sum_{k=1}^{2}\left[H\left(P_{S_j}^k, y_{S_j}\right) + H\left(P_S^f, y_{S_j}\right)\right], \quad (9)$$

where $H(\cdot)$ stands for the cross-entropy loss. Notably, the supervised loss $\mathcal{L}_{cls}$ is only calculated using labeled data from the source domain, and the fusion classifier $C^f$ in the feature fusion module is optimized with this loss exclusively.

### D. Training and Inference

During the training stage, the total objective function of our model is defined as follows:

$$\mathcal{L} = \alpha * \mathcal{L}_{dtl} + \beta * \mathcal{L}_{con} + \lambda * (\mathcal{L}_{mmd} + \mathcal{L}_{rest}) + \mathcal{L}_{cls}, \quad (10)$$

where the hyper-parameters $\alpha$, $\beta$, $\lambda$ are used to balance the weights of various losses. Notably, the training stage of BTMuda is summarized in Algorithm 1 for a better understanding. During the inference stage, two paths of the domain mixed feature extractor and all alignment sub-modules are involved. Since only images from the target domain are input to the network, we only utilize the target branch in the three-branch Transformer. After going through alignment sub-modules in two paths, we adopt the feature fusion module with $C^f$ instead of using source-specific classifiers $\{C_j^k\}_{k=1, j=1}^{2*M}$ to make a more comprehensive inference for our final prediction $P_{\mathcal{T}}^f$. The formula for the inference process is defined as follows:

$$P_{\mathcal{T}}^f = C^f\left(Cat_{j=1...M; k=1,2}\left(A_j^k\left(F^k(x^{\mathcal{T}})\right)\right)\right). \quad (11)$$

### E. Implementation Details

We implement our BTMuda with PyTorch on an NVIDIA GeForce GTX 3090 GPU. In the domain mixed feature extractor, the CNN path uses a ResNet-50, while the Transformer path uses the ViT-Small. For each dataset, the input size of all images is adjusted to 512*512. We apply data augmentation on both source and target domains during the training stage, including color jitters (such as brightness, contrast, and hue) and transformations (such as horizontal flipping, random rotation, and random resized cropping).

Due to the invisibility of labels in the target domain and the insufficiency of medical images during training, we use the BYOL [41] and MAE [42] to pre-train the BTMuda. BYOL injects contrastive learning into an asymmetric structure to encourage the student and teacher model to produce similar representations given differently augmented inputs, while MAE establishes a self-supervised mask reconstruction (SSMR) task where it randomly masks a large number of image patches in mammography images and reconstructs them in a self-supervised manner. We use different pretraining strategies to explore their gains. A detailed discussion of the gains of these pre-training strategies is in Section IV.D.

For training, we use the five-fold cross-validation method, where the batch size is set to 8 and the number of training iterations is 40000. Regarding the optimizer strategy, we employ the standard mini-batch stochastic gradient descent (SGD) algorithm with an initial learning rate of 0.001, a momentum of 0.9, and a weight decay of 0.0005. We also employ the annealing strategy from RevGrad [43] to adjust the learning rate. As for the hyper-parameters in Eq. 12, $\alpha$ and $\beta$ are used to balance the distillation loss and consistency loss. $\beta$ is dynamically given a weight using a Gaussian-like increasing function as $\beta = \beta_{max} * exp\left(-\delta\left(1 - \frac{e}{er}\right)^2\right)$, while $\alpha$ is always kept as 1 in the experiment. In this function, $\delta$ (set to 0.65) determines the shape of the Gaussian-like incremental function, and $\beta_{max}$ (set to 0.5) denotes the maximum value of $\beta$. With such a loss function strategy, the distillation loss can provide indispensable soft supervision to mitigate the effect of inter-domain shift throughout the whole training process. As the training progresses and $\beta$ gradually approaches $\beta_{max}$, the model is endowed with more feature extraction capability, allowing the consistency loss to play a greater role in dealing with the domain shift issues. We set the parameters $\lambda$ to control the losses in two alignment modules that have great importance. To suppress noise in the early stages of training, $\lambda$ is not fixed in the experiments but instead is set using the equation $\lambda = \frac{2}{exp(-\theta * p)} - 1$. $\lambda$ changes gradually from 0 to 1 asymptotically, while $\theta$ is fixed to 10 throughout the experiment. This asymptotic strategy greatly stabilizes the sensitivity of $\lambda$ and simplifies the choice of the model. All hyper-parameter selection experiments will be analyzed in detail in the ablation experiments.

## IV. EXPERIMENTS AND RESULTS

### A. Datasets and evaluation metrics

We evaluate our BTMuda on three public mammographic datasets: CBIS-DDSM [44], InBreast [2], and MIAS [45]. **CBIS-DDSM** is an updated and standardized version of the DDSM dataset, consisting of 2886 samples presented with 2D images in DICOM format. Each sample is diagnosed as normal/benign or malignant. According to the official instructions, we label the normal/benign samples as negative ones, and the malignant samples as positive ones in our experiments. Finally, we get 1429 negative images and 1457 positive images. **InBreast** contains 410 full-field digital images with BI-RAIDS readings. Following the instructions, we assign images with BI-RAIDS readings of 1, 2 as benign samples, and 4,5,6 as malignant samples, while the BI-RADIS reading of 3 is ignored since it has no clear designation of benign or malignant class. Same as CBIS-DDSM, the benign samples and the malignant samples are viewed as negative samples and positive samples, respectively. **MIAS** contains 322 scanned copies of films diagnosed with benign, malignant, or normal, and we label them in the same way as in CBIS-DDSM. All the images in these datasets will be converted into 2D PNG format before being input into the network.

For training, we fix CBIS-DDSM as the source domain because it has the most images, while we use MIAS and InBreast as the target and source domains, respectively, in different experiments. For evaluation, we adopt the most commonly used metrics in image classification. On the one hand, we compute the area under the curve (AUC) according to the receiver operator characteristic (ROC) curve. The ROC curve summarizes the tradeoff between the true-positive rate and false-positive rate for a model using different probability thresholds. On the other hand, we also use the F1 score, which is the harmonic mean of the precision and recall, as well as the accuracy (ACC) to evaluate the models.

### B. Compared methods

To verify the efficacy of our BTMuda, we compare it with multiple state-of-the-art methods. Specifically, the breast cancer classification method includes PHAM [36], the SUDA method includes MCD [13], while MUDA methods comprise: M³SDA [18], T-SVDNet [46]; DualMarker [15], DSFE [47], and TFFN [32]. For SUDA methods, two protocols are adopted: (1) Single Best, which reports the best result among all source domains, by comparing these results, we can evaluate whether we have improved the model's performance by introducing the multiple source domains or not and (2) Source Combined, which naively combines all source domains and then performs single-source domain adaptation. Source-Only refers to directly transferring the model trained in source domains to the target domain. To ensure a fair comparison, the results of these methods are obtained either from their respective papers or reimplemented based on their original papers or released code. We maintain consistency by employing the same backbone architecture and data pre-processing routines for all compared methods on each dataset.

TABLE I
COMPARISON WITH SoTA METHODS ON MIAS. BEST RESULTS ARE EMPHASIZED IN **BOLD** AND THE SECOND-BEST ONES ARE UNDERLINED.

| Protocols | Methods | ACC (%) | AUC | F1 |
|---|---|---|---|---|
| Single Best | Source-only | 62.31 | 0.554 | 0.609 |
|  | PHAM [36] | 65.09 | 0.557 | 0.625 |
|  | MCD [13] | 57.67 | 0.505 | 0.548 |
| Source Combine | Source-only | 63.35 | 0.560 | 0.626 |
|  | PHAM [35] | 66.16 | 0.580 | 0.683 |
|  | MCD [12] | 60.75 | 0.508 | 0.538 |
| Multi-Source | M³SDA [18] | 59.75 | 0.602 | 0.608 |
|  | T-SVDNet [46] | 60.67 | 0.614 | 0.632 |
|  | DualMarker [15] | 72.36 | 0.657 | 0.708 |
|  | DSFE [47] | 70.14 | 0.643 | 0.679 |
|  | TFFN [32] | 69.25 | 0.638 | 0.662 |
|  | Ours(average) | <u>78.26</u> | <u>0.689</u> | <u>0.774</u> |
|  | Ours (fusion) | **81.00** | **0.725** | **0.810** |

TABLE II
COMPARISON WITH SoTA METHODS ON INBREAST. BEST RESULTS ARE EMPHASIZED IN **BOLD** AND THE SECOND-BEST ONES ARE UNDERLINED.

| Protocols | Methods | ACC (%) | AUC | F1 |
|---|---|---|---|---|
| Single Best | Source-only | 52.45 | 0.525 | 0.361 |
|  | PHAM [36] | 64.89 | 0.515 | 0.604 |
|  | MCD [13] | 56.97 | 0.503 | 0.439 |
| Source Combine | Source-only | 50.72 | 0.551 | 0.383 |
|  | PHAM [35] | 66.95 | 0.543 | 0.588 |
|  | MCD [12] | 58.09 | 0.525 | 0.436 |
| Multi-Source | M³SDA [18] | 61.00 | 0.539 | 0.581 |
|  | T-SVDNet [46] | 59.21 | 0.545 | 0.553 |
|  | DualMarker [15] | 67.74 | 0.646 | 0.659 |
|  | DSFE [47] | 65.93 | 0.622 | 0.640 |
|  | TFFN [32] | 64.17 | 0.617 | 0.629 |
|  | Ours(average) | <u>71.03</u> | <u>0.653</u> | <u>0.668</u> |
|  | Ours (fusion) | **73.64** | **0.683** | **0.718** |

### C. Comparison with the State-of-the-art Methods

Table I illustrates the quantitative comparison results on the MIAS dataset. Generally, the results of Source Combination are better than those of Single Best, which demonstrates that it is feasible to improve the performance by combining all source domains into one domain. This may be due to the data enrichment with more sufficient training samples. In the multi-source protocol, we record not only the result derived from the proposed feature fusion module, i.e., BTMuda (fusion) but also the average result of all classifiers in our BTMuda, i.e., BTMuda (average). As can be seen, even if it is BTMuda (average), it can exceed the second-best method DualMarker by 5.9% ACC, 0.032 AUC, and 0.066 F1. Furthermore, our proposed BTMuda (fusion) lifts the performance to the highest level of 81.00% ACC, 0.725 AUC, and 0.810 F1, proving the effectiveness of our BTMuda in improving performance using multi-source data.

TABLE III
ABLATION STUDIES ON KEY COMPONENTS OF BTMUDA ON TWO DATASETS.

| Exp. ID | $F^C$ | $F^T$ | $\mathcal{L}_{mmd}$ | $\mathcal{L}_{rest}$ | three branch | $\mathcal{L}_{con}$ | Accuracy(%) MIAS | Accuracy(%) InBreast | AUC MIAS | AUC InBreast | F1 MIAS | F1 InBreast |
|---|---|---|---|---|---|---|---|---|---|---|---|---|
| I | √ | - | √ | - | - | - | 68.53 | 65.89 | 0.624 | 0.545 | 0.619 | 0.605 |
| II | √ | - | √ | √ | - | - | 69.47 | 66.92 | 0.629 | 0.569 | 0.623 | 0.645 |
| III | - | √ | √ | - | - | - | 69.15 | 66.40 | 0.600 | 0.555 | 0.599 | 0.613 |
| IV | - | √ | √ | √ | - | - | 70.40 | 67.44 | 0.617 | 0.573 | 0.617 | 0.647 |
| V | - | √ | √ | √ | √ | - | 75.39 | 68.99 | 0.646 | 0.609 | 0.639 | 0.682 |
| VI (baseline) | √ | √ | - | - | - | - | 62.31 | 52.45 | 0.624 | 0.565 | 0.609 | 0.361 |
| VII | √ | √ | √ | - | - | - | 72.90 | 67.18 | 0.631 | 0.577 | 0.635 | 0.629 |
| VIII | √ | √ | √ | √ | - | - | 74.45 | 68.47 | 0.652 | 0.601 | 0.679 | 0.654 |
| IX | √ | √ | √ | √ | √ | - | 76.32 | 69.77 | 0.678 | 0.649 | 0.744 | 0.685 |
| **X (proposed)** | √ | √ | √ | √ | √ | √ | **81.00** | **73.64** | **0.725** | **0.683** | **0.810** | **0.718** |

TABLE IV
RESULTS ON MIAS OF OUR BTMUDA, WHICH ADOPTS DIFFERENT PRE-TRAINING STRATEGIES FOR THE TWO PATHS OF THE DOMAIN MIXED FEATURE EXTRACTOR.

| Methods | | Accuracy(%) | AUC | F1 |
|---|---|---|---|---|
| ResNet-50 (CNN Path) | Three-branch Transformer (Transformer Path) | | | |
| without pre-training | training from scratch | 74.47 | 0.679 | 0.753 |
| without pre-training | pre-trained with SSMR | 71.31 | 0.686 | 0.738 |
| pre-trained on ImageNet | training from scratch | 75.69 | 0.678 | 0.764 |
| pre-trained on ImageNet | pre-trained with SSMR | 76.30 | 0.669 | 0.767 |
| ImageNet initialization and Self-supervised with BYOL | training from scratch | **81.00** | **0.725** | **0.810** |
| ImageNet initialization and Self-supervised with BYOL | pre-trained with SSMR | 71.90 | 0.657 | 0.685 |

Table II reports the results of adaptation on the InBreast dataset. The InBreast dataset has a higher resolution compared to the MIAS dataset, which can create a more severe intra-domain shift that brings more challenges for the model to distinguish between different samples within the same domain. This can be reflected by that all the results degrade to a lower level. Even so, the Source Combination protocol still performs better than the Single Best protocol. Our BTMuda (fusion) still ranks first and outperforms the second-best method DualMarker by 5.9% ACC, 0.037 AUC, and 0.059 F1, which demonstrates the superiority and robustness of our method.

*D. Ablation Study*

*(1) Evaluations on Key Components:* To evaluate the contributions of the key components in our method, we progressively conduct the ablation study. The quantitative results are summarized in Table III, from which several observations can be concluded:

1) Contribution of domain mixed feature extractor: The performance of our ultimate model (Exp. X) is significantly better than when only using the CNN (Exp. II) or Transformer (Exp. V). This suggests that both paths in the proposed domain mixed feature extractor are indispensable for extracting features with intra-domain robustness (i.e., having both global and local senses). Additionally, when we remove the consistency loss between the two paths (Exp. IX), there is a noticeable drop in accuracy of 4.68% on MIAS and 3.87% on InBreast. This demonstrates the importance of the consistency constraint, as it enforces consensus between the predictions from the two feature extraction paths and improves the robustness of the final prediction.

2) Contribution of two alignment modules: In Exp. VI, the network with bare two paths, also known as the "Source-only" in the previous section, is considered as the baseline. As expected, this baseline performs poorly because it only employs supervised learning on labeled source data without any alignment process. However, by simply introducing the source-target pair alignment module with the MMD loss (Exp. VII), we are surprised to find that the accuracy of the model is already better than some comparison methods. This shows that the feature alignment procedure plays a crucial role in aligning the source and target domains. Besides, adding the decision boundary module with the restriction loss (Exp. VIII) further improves the accuracy of the model by an average of 1.42% on two datasets. When the module is added to CNN and Transformer paths separately (Exp. II and Exp. IV), it also leads to significant improvements compared to Exp. I and Exp. III without the module. These comparisons demonstrate that the decision boundary alignment module can restrict all classifiers to maintain the same prediction on target domain samples, leading to more reliable predictions.

3) Contribution of three-branch architecture in the Transformer path: By upgrading from a vanilla Transformer (Exp. IV) to a three-branch one (Exp. V), we observe an average increase of 3.27%, 0.033, and 0.029 on three metrics in two datasets. Additionally, compared with Exp. VIII, the incorporation of a three-branch architecture in the co-trained CNN and Transformer (Exp. IX) also promotes the accuracy by 1.87% and 1.30% in MIAS and InBreast, respectively.

TABLE V
COMPARISON RESULTS OF DIFFERENT METHODS.

| Exp. ID | Methods | Params (M) | MACs (G) | Training time (h) | Inference time (s) | Accuracy (%) | AUC | F1 |
|---|---|---|---|---|---|---|---|---|
| I | Vanilla Resnet-50 [48] | 25.56 | 21.58 | 6.40 | 0.54 | 60.26 | 0.604 | 0.625 |
| II | Vanilla Vit-Small [49] | 22.00 | 24.04 | 6.55 | 0.56 | 61.54 | 0.590 | 0.578 |
| III | Vanilla Hybrid Model | 47.56 | 45.62 | 6.72 | 0.58 | 62.31 | 0.624 | 0.609 |
| IV | Only CNN Path | 26.77 | 22.08 | 6.45 | 0.57 | 69.47 | 0.629 | 0.623 |
| V | Only Transformer Path | 23.31 | 24.62 | 6.65 | 0.58 | 75.39 | 0.646 | 0.639 |
| VI | BTMuda w/o three-branch | 50.66 | 46.76 | 6.69 | 0.54 | 74.45 | 0.652 | 0.679 |
| VII | BTMuda (ours) | 50.76 | 46.88 | 6.75 | 0.53 | 81.00 | 0.725 | 0.810 |

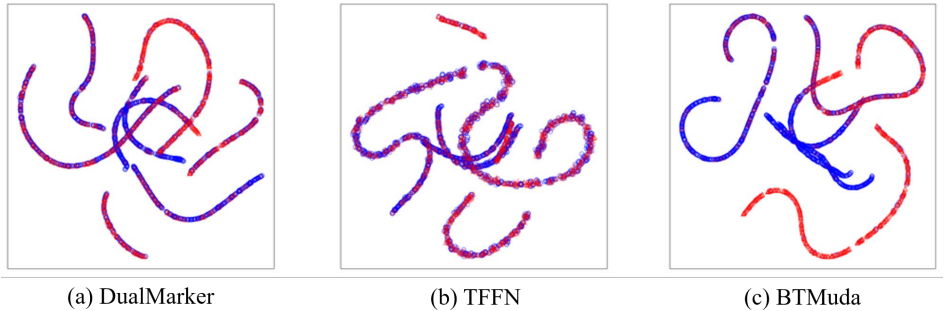

(a) DualMarker     (b) TFFN     (c) BTMuda

**Fig. 4.** T-SNE visualizations of feature embeddings for different models on the MIAS dataset. The red and blue points represent target samples with the positive and negative categories, respectively.

These findings suggest that leveraging the extra source-target branch in the three-branch Transformer will guide the target branch to learn more inter-domain invariant representations across domains through the use of three-branch cross-attention calculation as well as two-branch distillation loss optimization.

*(2) Exploration of Different Pre-training Strategies:* In this section, we study the effect of different pretraining strategies on the performance of our domain mixed feature extractor on mammography images. For the CNN path, we consider three pretraining strategies: (1) training from scratch (without pre-training), (2) pretraining on ImageNet, and (3) starting with ImageNet initialization followed by self-supervised learning (BYOL) on three medical image datasets. For the Transformer path, we do not use the existing ViT model pre-trained on ImageNet due to compatibility issues with our three-branch Transformer, and retraining our proposed Transformer using ImageNet would also require significant time and resources. Therefore, we consider two pretraining strategies for the Transformer path: (1) training from scratch and (2) self-supervised learning of mask reconstruction (SSMR) on three medical image datasets. In our experiments, BYOL [41] and SSMR [42] follow the experimental configurations suggested by their original papers, respectively.

The results of these pretraining strategies are shown in Table IV. We find that for the CNN path, models pre-trained on ImageNet in a supervised manner perform slightly better than those trained from scratch but worse than those pre-trained using self-supervised methods (BYOL) by a large margin. This may be because the distributions of the datasets used for the pretraining task and the downstream task are significantly different, which will compromise the improved performance of the pre-trained model. In contrast, self-supervised schemes construct some elaborate tasks to force the model to self-constrain, providing it with a more generalized and comprehensive understanding of the data distribution in medical domains. As for the transformer path, using excessively difficult pre-training tasks may be counterproductive, as indicated by the comparison of rows five and six in Table IV. As observed, both the CNN and Transformer paths are pre-trained in a self-supervised manner, but this combination results in an unexpected degradation in the model's performance. Among the six pretraining strategies for the two datasets, the best results for all three metrics are achieved using a CNN initialized with ImageNet and pre-trained with self-supervised BYOL, and a three-branch Transformer without pretraining.

*(3) Computational and Time Complexity Analysis:* To further verify the superiority of our method, we conduct the comparison of model size (e.g., Params), computational complexity (e.g., Multiply-Accumulate Operations, MACs), and time complexity (e.g., the total running time in both the training and inference phases) on the MIAS dataset. The experiments were performed on an 11th Gen Intel(R) Core(TM) i9-11900K @3.50GHz with 8 processors and an NVIDIA GeForce RTX 3090 GPU. Notably, the input size of all images is sized to 512x512 pixels, the term "Vanilla Hybrid Model" refers to the combination of Vanilla Resnet-50 [48] and Vanilla Vit-Small [49]. As observed from Table V, we can identify the following findings: (1) when juxtaposing our methods with Vanilla models (Exp. IV with Exp. I, Exp. V with Exp. II, and Exp. VII with Exp. III), respectively. Our methods consistently demonstrate impressive performance while maintaining relatively small increases in model size, computational complexity, and training time. Concretely, when comparing BTMuda with the Vanilla Hybrid Model, the increase of model parameters (2.7M), MACs (1.26G), and training time (0.03h) is marginal, while the performance improvements of 18.69% for accuracy, 0.101 for AUC, and

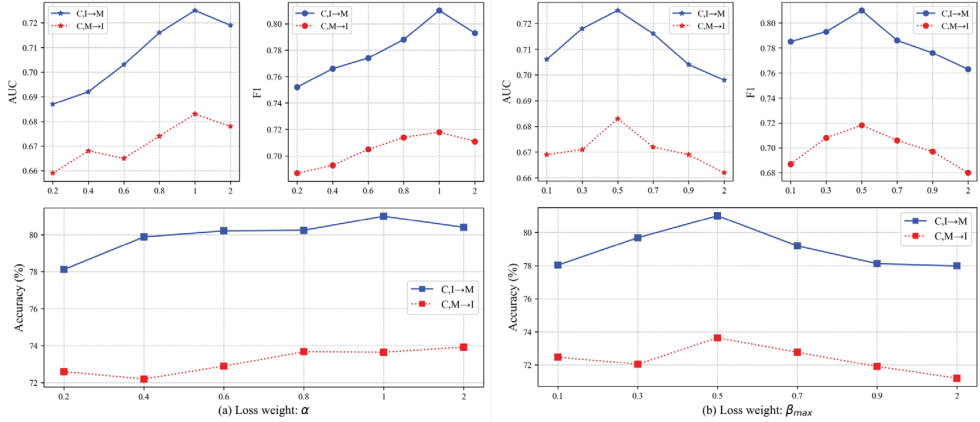

**Fig. 5.** Parameter sensitivity tests for $\alpha$, $\beta_{max}$ on the MIAS and InBreast.

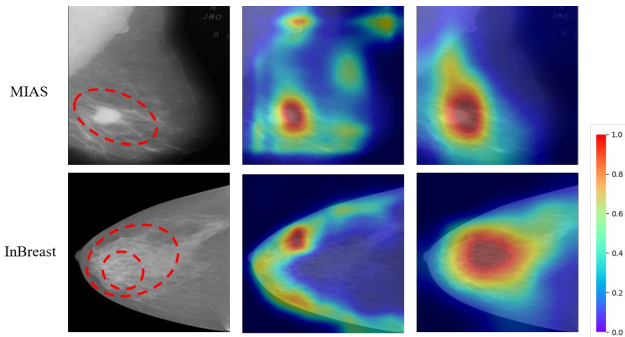

**Fig. 6.** The Grad-CAM visualizations of our proposed method BTMuda (third-column) and the compared method TFFN (second-column) on the MIAS and InBreast datasets. Each row contains an original image and its corresponding class activation map, which highlights the regions of the image that contribute most to the model's prediction. These regions are indicated by red circles in the original images.

0.201 for F1 are significant; (2) the strategic integration of our three-branch architecture in the Transformer path yields improvements across all three metrics without substantial increases in model size, computational complexity, or training time, as evidenced by the comparative analyses between Exp. VII and VI. Overall, these experimental results demonstrate that our method can achieve notable performance enhancement with nearly the same resources.

*(4) Visualization of Feature Embeddings:* To demonstrate the effectiveness of our model in transferring knowledge from the source domain to the target domain, we visualize the feature embeddings of different models on the MIAS dataset using t-SNE visualizations. As shown in Fig. 4, for DualMarker [15] and TFFN [32], the two classes in the target domain are almost indistinguishable and overlap significantly, making it difficult to differentiate and produce reliable predictions. In contrast, the visualization produced by our BTMuda shows clear boundaries between clusters, indicating that it has a better transfer ability and can eliminate domain discrepancy without sacrificing discrimination ability. Besides, the two classes are more clearly separated in the visualization of our BTMuda, with each class forming a distinct cluster at opposite ends of the visualization. Overall, these results demonstrate the superiority of our BTMuda qualitatively.

*(5) Hyper-parameters Sensitivity:* In this section, we conduct a series of experiments on MIAS and InBreast to evaluate the values of the manually set hyper-parameters in Eq. 12. The hyperparameter $\alpha$ controls the importance of the distillation loss ($\mathcal{L}_{dtl}$), while $\beta$ relied on the $\beta_{max}$ and controls the importance of the consistency loss ($\mathcal{L}_{con}$). First, to study the sensitivity of $\alpha$, we sample its values in {0.2, 0.4, 0.6, 0.8, 1, 2}, and fix $\beta_{max}$=1. As can be seen in Fig. 5 (a), the performance fluctuates within a relatively small range with $\alpha$ changing, demonstrating that BTMuda is robust to the distillation loss weight $\alpha$ and our method gains the best performance when $\alpha$ is set to 1. Then we set $\alpha$ to 1, and adjust $\beta_{max}$ from 0.1 to 2. Fig. 5 (b) shows that when $\beta_{max}$ is set to 0.5, the proposed method obtains the highest score in all three metrics. Either less than 0.5 or more than 0.5 leads to performance degradation, as the interaction of information between the two feature extractors is necessary, but too much interaction may cause interference. Therefore, we determine $\beta_{max}$ as 0.5. Based on these results, we set $\alpha$=1, $\beta_{max}$=0.5 to achieve the best performance of our BTMuda.

*(6) Visualization of Class Activation Map:* Fig. 6 illustrates some examples from the MIAS and InBreast datasets, along with the Grad-CAM [50] visualizations of the proposed BTMuda model and the compared method TFFN [32]. The Grad-CAM technique generates class activation maps that highlight the regions of the image that contribute most to the model's prediction. These regions, depicted as red circles in the visualizations of Fig. 6, are frequently associated with abnormalities in the breast tissue, such as masses and calcifications, which may indicate the presence of breast cancer [2]. As shown in Fig. 6, the TFFN exhibits a propensity for distraction by irrelevant details on the MIAS dataset, particularly misplaced focus on the non-informative background of the images. In contrast, our proposed BTMuda method maintains a robust capacity to concentrate on the dense regions of breast tissue across the two datasets, which demonstrates the proficiency of BTMuda to accurately detect breast cancer. On the Inbreast dataset, although not perfectly attuned to the primary tumor areas of breast tissue, our model is more sensitive in targeting dense areas, which is crucial for the clinical detection of early-stage breast cancer and assists physicians in making further clinical diagnoses. Overall, our proposed BTMuda method maintains a robust capacity to concentrate on the dense regions of breast tissue across the two datasets, demonstrating the proficiency of BTMuda to accurately detect breast cancer.

## V. Discussion

In this paper, we proposed BTMuda, a novel multi-source domain adaptation framework developed for breast cancer screening using 2D mammography data. Our method effectively addresses the challenges of insufficient data and domain shift in multi-source medical data analysis by dividing the domain shift issue into two levels: intra-domain and inter-domain. While BTMuda has achieved good performance compared to other MUDA methods, there is still room for improvement compared to supervised methods. This is a common challenge in MUDA, as the lack of labeled data in the target domain can sometimes lead to the learning of useless or even wrong knowledge. However, we are actively working on ways to improve BTMuda by extracting more useful domain-invariant representations from multiple sources and target domains and applying more effective complementary supervision. Furthermore, in light of the scarcity of 3D datasets and the absence of artificially labeled segmentation masks in certain datasets, such as MIAS [45], our study predominantly concentrates on the classification task utilizing 2D data, which is aligned with the traditional breast cancer diagnosis works [11, 36]. In future work, we are committed to expanding our research by amassing an extensive collection of both 2D [51] and 3D [52] datasets with segmentation masks, aiming to facilitate the investigation of more pragmatic and clinically relevant tasks. Additionally, we are also aware of the complexity of hyper-parameter settings when using multiple modules and the potential risk of model instability. In the future, we will address these challenges by considering the use of techniques such as mutual learning to strengthen consensus among modules or adopt stronger methods for handling hard samples. Despite these limitations, we believe that BTMuda has the potential to make a significant impact in the field of breast cancer screening using mammography.

## VI. Conclusion

In this paper, we presented a novel multi-source domain adaptation framework called BTMuda for breast cancer screening using 2D mammography. Our approach addresses the challenges of insufficient data and domain shift in multi-source medical datasets by dividing the domain shift issue into bi-level: intra-domain and inter-domain. Concretely, to reduce the intra-domain shift, we proposed a domain mixed feature extractor consisting of a CNN and a Transformer to extract both local and global features, which allows the model to have a more nuanced and comprehensive understanding of each domain. To address the inter-domain shift, we redesigned the Transformer with cross-attention and distillation to extract inter-domain invariant representations from multiple sources. In addition, we introduced two alignment modules, one for feature alignment and one for prediction alignment, to further fine-tune the alignment process. By effectively addressing the bi-level domain shifts, our approach yields robust results and outperforms previous methods by a large margin on three public mammographic datasets. This demonstrates the potential of BTMuda and other MUDA approaches in the medical analysis community, and we hope that our work will inspire future research in this area.

Acknowledgment Statement

All authors declare that they have no known conflicts of interest in terms of competing financial interests or personal relationships that could have an influence or are relevant to the work reported in this paper.